\documentclass[journal]{IEEEtran}
\usepackage{amsmath}
\usepackage{amssymb}
\usepackage{cite}
\usepackage{array}
\usepackage{multirow}
\usepackage{float}
\usepackage{algorithm}
\usepackage{algorithmic}
\usepackage{caption}
\usepackage{bm}
\usepackage{booktabs}
\usepackage{graphicx}
\usepackage{longtable,tabularx}
\usepackage{subfigure}
\ifCLASSINFOpdf
\else
\fi
\begin{document}
%
\title{Mission schedule of agile satellites based on Proximal Policy Optimization Algorithm}
%
%
%

\author{Xinrui Liu, xinrui.loveliu@gmail.com,https://github.com/loceyi/satellitesMissionSchedule}

%
%

\markboth{Mission schedule of agile satellites based on Proximal Policy Optimization Algorithm}%
{Shell \MakeLowercase{\textit{et al.}}: Bare Demo of IEEEtran.cls for IEEE Journals}
%



\maketitle

\begin{abstract}
Mission schedule of satellites is an important part of space operation nowadays, since the number and types of satellites in orbit are increasing tremendously and their corresponding tasks are also becoming more and more complicated. In this paper, a mission schedule model combined with Proximal Policy Optimization Algorithm(PPO) is proposed. Different from the traditional heuristic planning method, this paper incorporate reinforcement learning algorithms into it and find a new way to describe the problem. Several constraints including data download are considered in this paper.
\end{abstract}

\begin{IEEEkeywords}
Mission schedule, planning
\end{IEEEkeywords}

%
\IEEEpeerreviewmaketitle

\section{Introduction}
%
%
%
%
\IEEEPARstart{N}{owdays} lots of countries and companies have successively proposed satellite constellation plans for various purpose, such as communication, remote sensing and military use. This situation makes mission planing of satellites an important topic. Due to the high price of satellite construction and operation, missions need to be reasonably allocated to satellites to maximize the benefits of operation. This topic is always related to operation research. The nature of satellite mission planning is a non-linear programming problem, which is also a NP-complete proble.

The basic model of satellites mission schedule is raised by Hall and Magazine(1994)\cite{hallMaximizingValueSpace1994a}. It built a model which is a integer linear programming problem(ILP). For single satellite, Verfaille et al(1996).\cite{verfaillieRussianDollSearch1996} used russian doll algorithm to improve the solution search with depth-first search brand and bound.Hall and Magazine(1994) raised dynamic programming algorithm with Lagrangian relaxation method. Gabrel and Vanderpooten(2002)\cite{gabrelEnumerationInteractiveSelection2002}
 used acyclic graph to solve the problem. Frank et al.(2001) \cite{frankPlanningSchedulingFleets2001} brought out greed hill-climbing search with stochastic variations for one satellite mission planning. Vasquez and Hao(2001)\cite{vasquezLogicConstrainedKnapsackFormulation2001} come up with Tabu search with knapsack formulation. Globus et.al(2003)\cite{globusSchedulingEarthObserving2003}
introduced evolutionary algorithm into the problem including genetic algorithm, simulated annealing, stochastic hill-climbing, and iterated sampling.It compares the results of different algorithms and concluded that simulated was the best.

For multi-satellites, Globus et al.(2004) \cite{globusComparisonTechniquesScheduling2004b}
demonstrated the pros and cons of different algorithms such as genetic algorithm and simulated annealing. Bianchessi et al.(2007)\cite{bianchessiHeuristicMultisatelliteMultiorbit2007} used tabu search to solve the schedule problem for two satellites with column generation method. Hwang et al.(2010)\cite{hwangMultiObjectiveOptimizationMultiSatellite2010} simplified the multi-satellites problem by limited number of single-orbit scheduling problems and solved it by genetic algorithm. Nelson(2012)
\cite{nelsonSchedulingOptimizationImagery} considered satellite constellation and two orbits. Column generation method is used in his thesis. Wu et al. (2013)\cite{wuAdaptiveSimulatedAnnealingbased2014} considered four-satellite constellation, multiple orbits and semi-agile satellite. That paper adopts adaptive simulated annealing with dynamic clustering. Xiaolu et al.(2014)\cite{xiaoluMultiSatellitesScheduling2014} took decomposition algorithm to find the optimal solution better.

As for the application of reinforcement learning in satellites mission scheduling, Wang et al.(2011) \cite{AlgorithmCooperativeMultiple} applied multi-agent reinforcement learning algorithm in the multi-satellite cooperative planning problem. It used a blackboard architecture to deliver the joint punishment operator in order to get the results which satisfied the constraints those missions required. H. Wang(2019) \cite{wangOnlineSchedulingImage2019} combined the problem with Dynamic and stochastic Knapsack Problem. The paper used deep reinforcement learning with A3C algorithm(Asynchronous Advantage Actor-Critic) and applied neural network in the learning process. However, no paper clearly articulates the framework for the use of reinforcement learning in satellite mission planning, which is an important purpose of this paper.


\section{Problem Description}
A satellites with a optimal sensor is considered in this paper. Different point targets are set as tasks that need to be selected by the satellite. The satellite needs to choose among those tasks and maximize the rewards of executing them. It's worth noting that the requirements and types of tasks are varied according to the actual needs. Thus, different constraints need to be considered such as time window, energy limit, storage limit. In this paper, the main purpose is to establish a framework for task planning using reinforcement learning so only basic constraints are considered for demonstrating the model more clearly.


\subsection{Basic Prerequisite}

\begin{figure}[hbt!]
\centering
\includegraphics[width=.5\textwidth]{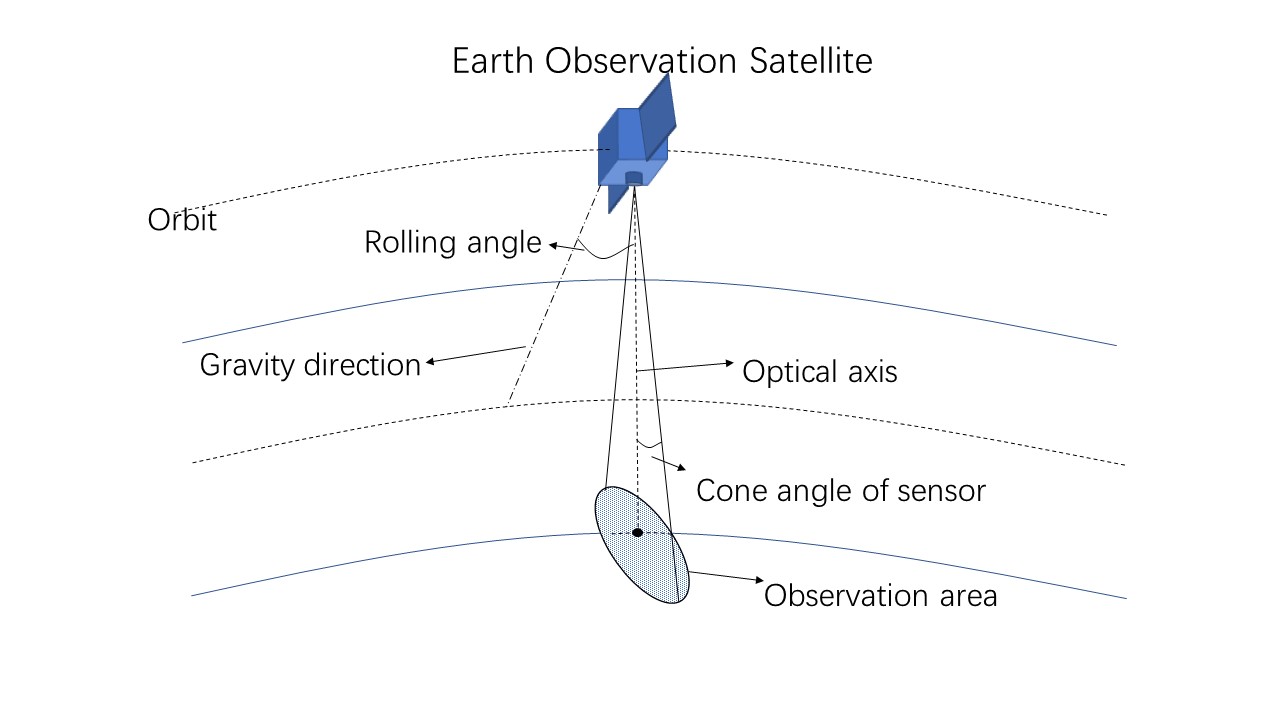}
\caption{Schematic diagram of satellite imaging process.}
\label{satellite}
\end{figure}

\begin{itemize}
\item \textit{Basic information of the planned satellite}

Only a single satellite is considered in this paper. It has fixed orbit and limited storage.

\item \textit{Optical sensor}

A optical sensor is installed on the satellite, which has a limited observation angle, as shown in Fig. \ref{satellite}. For better observation, the ground track of the sensor's optical axis of  is required to pass through the point target. As not considering the pitch angle, the corresponding roll angle is
the angle between the optical axis and the direction of gravity when point target is contained on the normal plane of satellites orbit direction.
\item \textit{No task interruption}

One satellite can only execute one task at a time. Also, a task can only be executed for one time. Once a task is received, no interruption is allowed in the execution process.

\item \textit{Award engineering}

Each task is assigned a reward. Once a task is completed, the corresponding reward will be given. The rewards are set according to their importance, in order to make important tasks more likely to be selected by the satellite. The final goal of our algorithm is to maximize the total reward.

\item \textit{Roll angle}

The agile satellite can swing to the left and right in orbit. The roll angle for each task is shown in Fig.\ref{satellite}. The corresponding roll angle is the angle between the optical axis and the direction of gravity when point target is contained on the normal plane of satellites orbit direction.

\item \textit{Simplified model}

The pitch angle is not considered in this paper since the main purpose is to clarify how to apply PPO algorithm to the mission plan, instead of building an accurate model to describe the satellite.

\end{itemize}
;

\section{Problem Formulation}
In order to solve the mission schedule problem by PPO, different equations need to be established to describe constraints. The schedule problem can be described as follows.
\subsection{Variable description}

\subsubsection{Preprocessing variables}
\begin{table}[H]
\renewcommand\arraystretch{2}
\renewcommand\tabcolsep{20.0pt}
\begin{tabular}{lp{6cm}}
 \textbf{Task}& Set of tasks $Task=[Task_1, Task_2, Task_3,...,Task_n]$ with $Task_i \in Task$.   \\
 $\textbf{TW}^T_i$ &  Time window of $Task_i$. It consists of  $\textbf{TW}^{T(s)}_i$ and $\textbf{TW}^{T(e)}_i$. In this paper, geostationary satellites are not considered. Thus, the relative position of the satellite and the task changes with time. The time period that the sensor can cover the target point is called the time window of the task. Time window is also considered as the required time consumption of those tasks.
\end{tabular}
\end{table}
\begin{table}[H]
\renewcommand\arraystretch{2.5}
\renewcommand\tabcolsep{20.0pt}
\begin{tabular}{lp{6cm}}
 $\textbf{TW}^S$  &  All the available time windows of the satellite. When a task is accepted, the available time should exclude the occupied time window of that task.\\
 $\bm{\varphi}_{i}$   & Required roll angle of $Task_i$.\\
 $\textbf{T}^{AM}_{ij}$   & Transition time between $Task_i$ and $Task_j$ for attitude maneuvers.\\
$\textbf{M}_{i}$   & Consumption of storage(memory) for $Task_i$. In this paper, the date download task is also considered.

$M_i=
\begin{cases}
\mbox{positive number} & \mbox{if $Task_i$ is an observation task}\\
\mbox{negative number} & \mbox{if $Task_i$ is a download task}
\end{cases}$
\\
$\textbf{M}^{S}$   & Maxium storage capability of the satellite. \\
$\bm{\varphi}_{max}$   & Maxium roll angle of the satellite. \\
 $\textbf{R}_{i}$   & Reward assigned to $Task_i$
\end{tabular}
\end{table}

\subsubsection{Pending variable}
\textbf{Schedule variables}
\begin{table}[H]
\renewcommand\arraystretch{3}
\renewcommand\tabcolsep{25.0pt}
\begin{tabular}{lp{6cm}}
$\textbf{D}_i$   &   Decision variables.
$D_i=
\begin{cases}
0 & \mbox{if $Task_i$ is refused}\\
1 & \mbox{if $Task_i$ is accepted}
\end{cases}$ \\
$\textbf{R}^T$ &  Total rewards  $R^T=\sum_{i\in Task} (D_{i} \cdot R_{i})$. The goal of our algorithm is to maximize it. The larger it is, the more successful the algorithm is. Hence it is an important index to evaluate the performance of the algorithm.
\end{tabular}
\end{table}

\textbf{Support variables}
\begin{table}[H]
\renewcommand\arraystretch{2}
\renewcommand\tabcolsep{25.0pt}
\begin{tabular}{lp{6cm}}
$\textbf{$\theta$}^{T}_{ij}$ &  Used to describe the execution sequence of tasks. If both $Task_i$ and $Task_j$ are
chosen, then \qquad \qquad \qquad
$\theta^{T}_{ij}=
\begin{cases}
1 & \mbox{if $Task_j$ is executed immediately after $Task_i$}\\
0 & \mbox{if not}
\end{cases}$\\
$\textbf{$\Gamma$}^{i}_{j}$ & \leftline{Describe the execution order in temporal relationship}
$\Gamma^{i}_{j}=
\begin{cases}
1 & \mbox{if $Task_j$ is the $i$th task in the timeline }\\
0 & \mbox{if not}
\end{cases}$
\end{tabular}
\end{table}


\subsection{Constraints}
Essentially, satellites mission planning is a constraint satisfaction problem(CSP). The goal is to find the values of schedule variables which satisfy the constraints we build and also try to maximize the reward $R_T$. The following content describe the constraints, which is derived from temporal and physical model of the satellite.

\subsubsection{Time window constraints}

\begin{equation}
\label{Constraints:Time window}
D_i \cdot \theta_{ij} \cdot TW^{T(e)}_i \leq D_j \cdot \theta_{ij} \cdot TW^{T(s)}_j, i\neq j
\end{equation}

Eq.\ref{Constraints:Time window} illustrates that time windows between selected tasks can not overlap beacause the satellite can only execute one task at a time.

\subsubsection{Roll angle constraints}

\begin{equation}
\label{Constraints:Roll angle}
\theta_{ij} \cdot (D_j \cdot TW^{T(e)}_{j} - D_i \cdot TW_i^{T(s)}) \geq T_{ij}^{AM}
\end{equation}
\begin{equation}
\label{Constraints:Roll angle2}
\varphi_{i} \leq \varphi_{max}, \forall i \in Task
\end{equation}

Eq.\ref{Constraints:Roll angle} shows that if $Task_i$ and $Task_j$ are arranged next to each other,
 the satellite must have sufficient time to maneuver from the roll angle of current task to the next task. Also, due to  maneuverability limitations of the satellite, the roll angle cannot exceed the maximum roll angle.

\subsubsection{Storage constraints}

\begin{equation}
\label{Constraints:Storage1}
T_{j_1}^{1} \cdot M_{j_1}\leq M^{S},  \forall j_1 \in Task
\end{equation}

\begin{equation}
\label{Constraints:Storage2}
T_{j_1}^{1} \cdot M_{j_1}+T_{j_2}^{2} \cdot M_{j_2} \leq M^{S},   \forall j_1, j_2 \in Task  \nonumber
\end{equation}
\begin{equation}
\label{Constraints:Storage3}
\vdots \nonumber
\end{equation}
\begin{equation}
\label{Constraints:Storage4}
\begin{split}
T_{j_1}^{1} \cdot M_{j_1}+T_{j_2}^{2} \cdot M_{j_2} +\cdots+ T_{j_n}^{n} \cdot M_{j_n} \leq M^{S},\\
 \forall j_1, j_2,..., j_n  \in Task, n = \sum_{i \in Task}D_i  \nonumber
\end{split}
\end{equation}

As shown in Eq.\ref{Constraints:Storage1}, the data generated by a task cannot exceed the remaining capacity of the satellite. Since data download is taken into consideration, we can not only limit all the data generated to be less than the storage capacity or we will lose the advantage that data download yields. This constraint needs to be considered by temporal sequence. Once a satellite begins its first task, the satellite should meet storage constraints at any time during the schedule horizon.

\section{PPO based model for schedule problem}
\subsection{Introduction of reinforcement learning}
\begin{figure}[h]
	\centering  
	\includegraphics[width=0.5\textwidth]{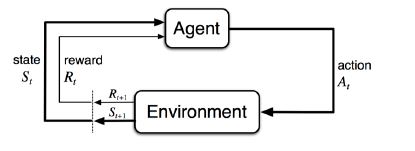}  
	\caption{The agent-environment feedback loop [Sutton and Barto, 1998]}  
	\label{fig:RL}   
\end{figure}

Reinforcement learning is learning what to do--how to map situations to action--so as to maximize a numerical reward signal.The learner is known
as the agent and everything outside the agent is known as the environment. The
agent selects actions and the environment responds by presenting a reward and a
new state. During the process of trial and error, the agent will learn from the experience and start to take the 'better' action in the state. A canonical view of this feedback loop is shown in Figure \ref{fig:RL}. The agent takes actions under current state and the environment will update the agent's state and return a reward as a feedback for the action.The agent keeps doing this until the environment gives back an end state and then a new episode could begin. During the interaction with environment, the agent gradually learns how to pick up the better action in each state and strive to maximize the final reward of next episode.

Besides the agent and the environment, there is another way to describe the learning system: a policy, a reward function, a value function.

Policy means the mapping from state to aciton, usually expressed as $\pi (S|A)$. It means the probability of taking action A under state S. The policy defines how the agent interact with the environment and a better policy will result in more rewards during the interaction.

Reward function is the mapping from state-action pair to a single number, a reward, which is designed artificially to guide the agent to find the best policy. It is a feedback from the environment and usually expressed as $R(S,A)$, which represents the reward when the agent takes action A under State S.

Value function is the mapping from a state to a number, which measures how good the state is. The ideal expression of value function is:

\begin{align}
\label{Value functions:1}
V^{\pi}(s)&=\mathbb{E}_{\pi}[R_t|s_t=s] \nonumber \\
&=\mathbb{E}_{\pi}[\sum^{\infty}_{k=0}\gamma^{k}r_{t+k+1}|s_t=s]\\ \nonumber
&=\sum_{a}\pi(s,a)\sum_{s'}P^a_{ss'}[R^a_{ss'}+\gamma V^{\pi}(s')]
\end{align}

\begin{figure}[H]
	\centering  
	\includegraphics[width=0.5\textwidth]{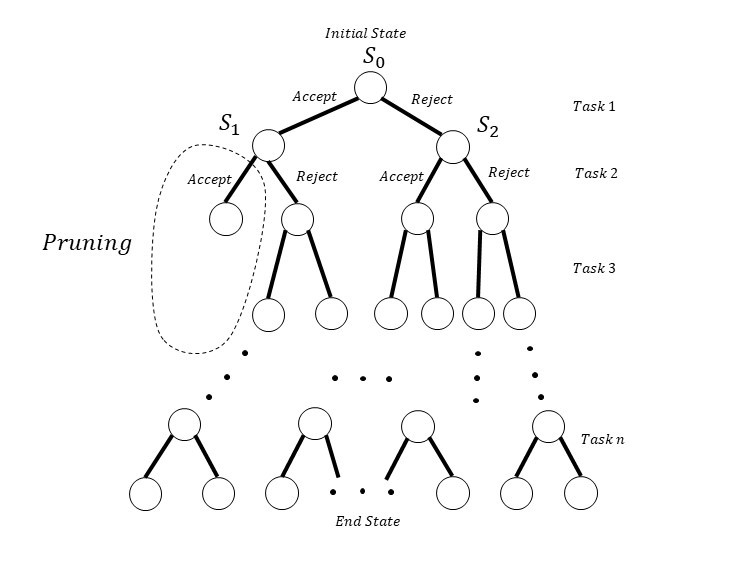}  
	\caption{Basic search process with pruning}  
	\label{fig:Pruning}   
\end{figure}

$P^{a}_{ss'}$ is the state transition probability from $s$ to $s'$. Normally, the interaction between the agent and the environment is considered a Markov Decision Process, which means taking the same action in the same state will lead to the transition to other states under different probability $P(s_i,a,s_{i+1})$. However, in the task schedule problem, the environment is set to be deterministic thus the next state obtained by taking the same action under the same state is fixed($P^{a}_{ss'}=1$). For our model, reinforcement learning method is no longer used as a way to learn how to interact with uncertain environment, on the contrary, it works as a search algorithm with evaluation of collected state and actions. The basic search process is shown in Fig.\ref{fig:Pruning}. As we can see, the agent constantly makes decisions about incoming tasks and the set of all actions made in each episode represent a solution for the schedule problem. Rather than trying to train an agent which can make a good schedule of tasks, the episode with maximum rewards during the training process is considered the best solution. A pruning trick is also applied in the search process, which will be elaborated in \ref{section:Model}

\subsection{PPO algorithm}

\begin{figure}[H]
	\centering  
	\includegraphics[width=0.5\textwidth]{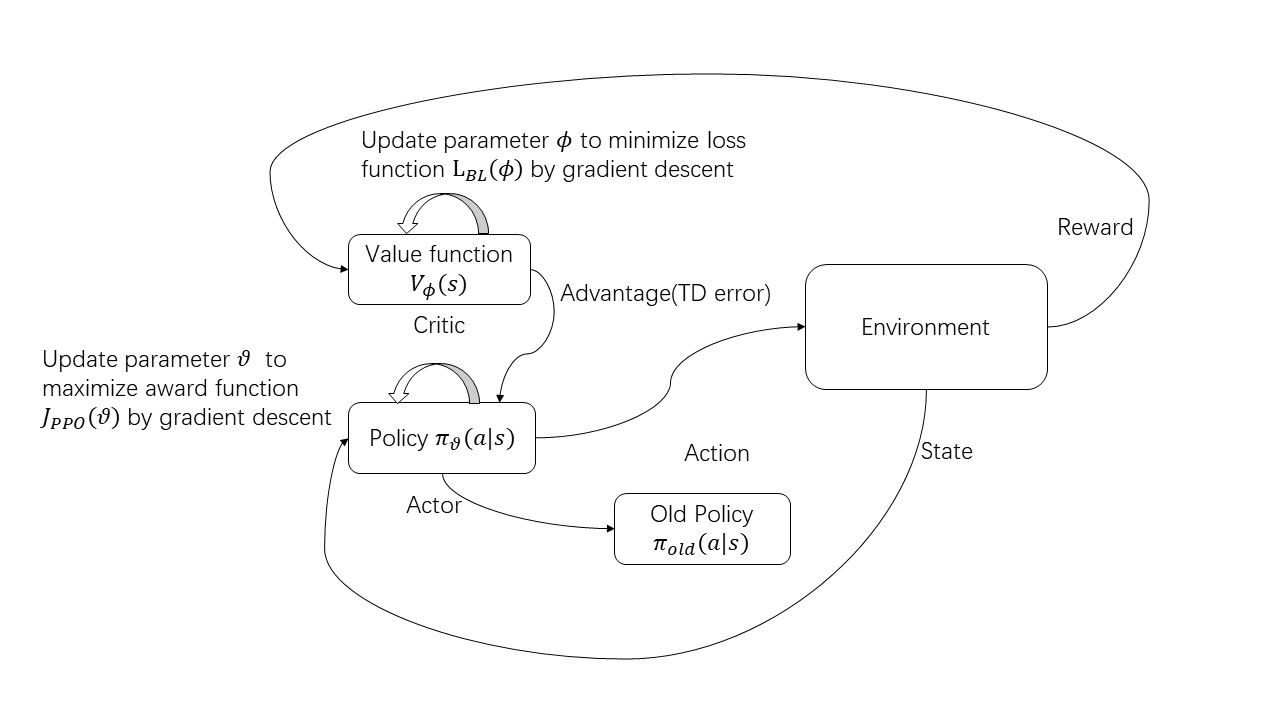}  
	\caption{Basic structure of PPO}  
	\label{fig:PPO}   
\end{figure}

PPO was the improved algorithm of Trust Region Policy Optimization raised by OpenAI\cite{schulmanTrustRegionPolicy2017}. It was published by both DeepMind and OpenAI in 2017\cite{heessEmergenceLocomotionBehaviours2017}
\cite{schulmanProximalPolicyOptimization2017}. PPO is based on policy gradient algorithm and actor-critic algorithm which is raised by G.Barto in 1983\cite{bartoNeuronlikeAdaptiveElements1983a}.
The structure of PPO are shown in Fig.\ref{fig:PPO}

\begin{figure}[H]
\centering
\subfigure[Neural Network(NN) of Critic(Value Function)] {\includegraphics[width=3.1in]{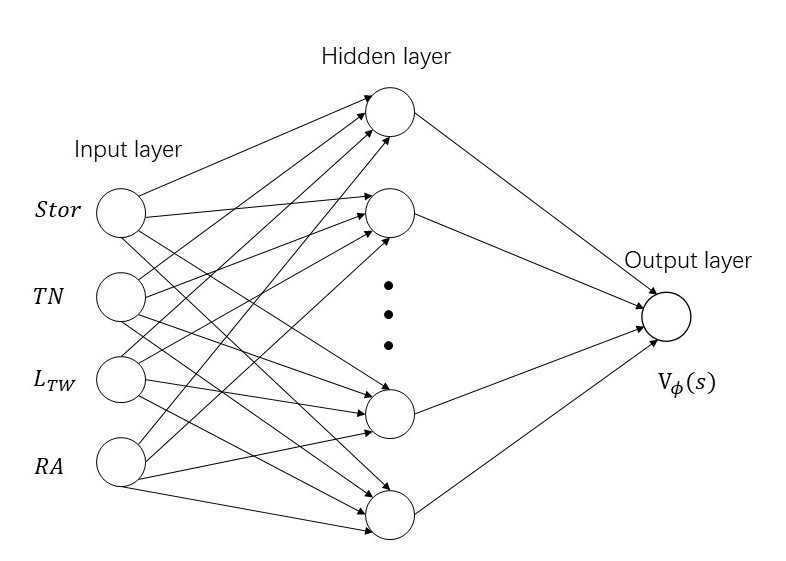}}
\subfigure[Neural Network(NN) of Actor(Policy)] {\includegraphics[width=3.3in]{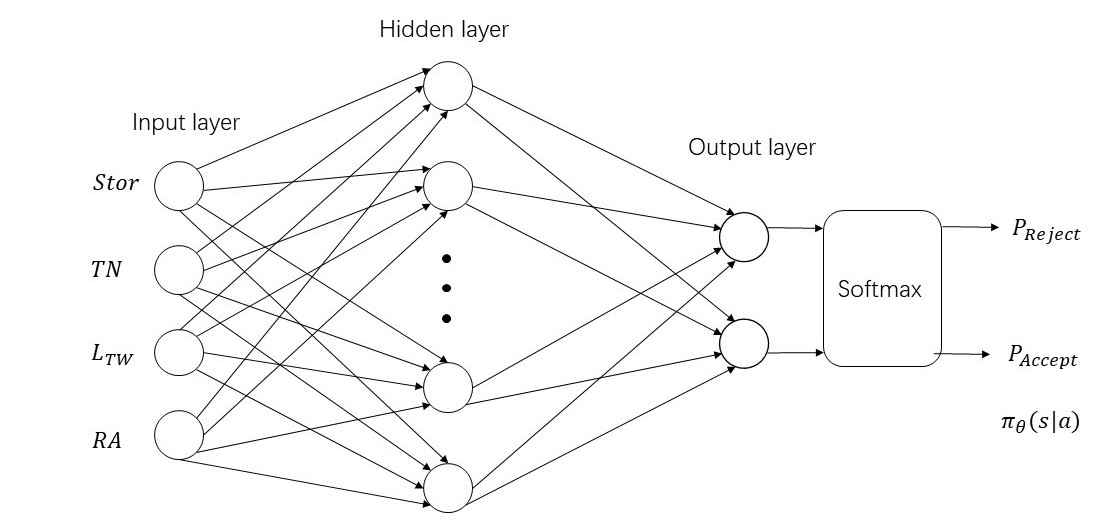}}
\caption{Neural Networks}
\label{fig:NN}
\end{figure}

There are three neural networks in PPO: new policy network(Actor), old policy network and value function network(Critic). The old policy network is just the copy of the new policy network to store the old parameters. The actor $\pi_{\theta}(a|s)$ is parameterized by $\theta$ and updated towards getting higher and higher total rewards. The critic parameterized by $\phi$ is updated to make output more close to the real situation. All of them consist of three fully connected layers as shown in Fig.\ref{fig:NN}


The activation function of the hidden layer in both NNs  is Relu activation function:
\begin{equation}
\label{ActivationFunction:Relu}
Relu(x)=
\begin{cases}
x & x \geq 0\\
0 & x < 0
\end{cases}
\end{equation}
Since the output of the actor's NN is the probability of selecting each action in a certain state, softmax function is used to transfer the results actor's output layer to the probability distribution. Softmax is a function as described in Eq.\ref{ActivationFunction:Softmax}. It takes as input a vector of K dimensions, and normalizes it into a probability distribution consisting of K probabilities.

\begin{equation}
\label{ActivationFunction:Softmax}
P_i=\frac{e^i}{\sum_{j}e^{j}}
\end{equation}
All of those networks' parameters are updated by gradient descent method toward maximize or minimize its loss function.

The pseudocode is shown below to illustrate the process better. The agent starts to interact with the environment based on the actor network until it goes into the end state. After that, the trajectory of the agent is used to update the parameters of actor-critc network. When updating the actor network, KL Penalty is introduced in the loss function. It is an index to evaluate the difference between $\pi_{old}$ and $\pi_{\theta}$. By adding it in the loss function, drastical change of the actor network will not happen which increases reliability of learning. For critic network, the goal of its update is to have a better evaluation of each state by summarizing the real rewards. After learning one episode, the agent will initialize its own state and begin next episode until meeting the requirement of maxium episode.

\begin{algorithm}[H]
\caption{Proximal Policy Optimization\cite{heessEmergenceLocomotionBehaviours2017}}
\begin{algorithmic}[1]
  \FOR{$i \in \left\{1,...,N \right\}$}
  \STATE Run policy $\pi_{\theta}$ until entering the end state, collecting $\left\{s_t, a_t, r_t  \right\}$
  \STATE Estimate advantages
  $\hat{A_t}=\sum_{t'>t}\gamma^{t'-t}r_{t'}-V_{\phi}(s_t)$
  \STATE  $\pi_{old} \leftarrow \pi_{\theta}$
  \FOR{$j \ in \left\{1,...,M\right\}$(M is the set number of Actor's update times)}
  \STATE $J_{PPO}(\theta)=\sum_{t=1}^{T}\frac{\pi_{\theta}
  (a_t|s_t)}{\pi_{old}(a_t|s_t)}\hat{A}_t-\lambda KL[\pi_{old}|\pi_{\theta}]$
  \STATE Update $\theta$ by a gradient method w.r.t $J_{PPO}(\theta)$
  \ENDFOR
  \FOR{$j \ in \left\{1,...,B\right\}$(B is the set number of Critic's update times)}
  \STATE $L_{BL}(\phi)=-\sum_{t=1}^{T}(\sum_{t'>t}\gamma^{t'-t}r_{t'}-V_{\phi}(s_t))^2$
  \STATE Update $\phi$ by a gradient method w.r.t $L_{BL}(\phi)$
  \ENDFOR
	\ENDFOR
	\end{algorithmic}
\end{algorithm}

\subsection{Scheduling model based on PPO}
\label{section:Model}
The satellite is the agent here. A virtual environment is built for interaction with the agent. This environment has its own rules for the interaction. The reward mechanism is also set artificially.

The satellite's state and action is defined below:
\begin{equation}
\label{State:definition}
S=\left\{Stor, TN, L_{TW}, RA\right\}
\end{equation}
\begin{equation}
\label{Action:definition}
A=\left\{0(Reject), 1(Accept)\right\}
\end{equation}

$Stor$ is the satellite's remaining storage capacity. When a task is selected, the corresponding storage state of the satellite needs to be changed according to the storage consumption of the task.

$TN$ is the serial number of the incoming task for which the satellite needs to make a decision(take an action). In our model, we reorder the tasks according to their execution time and provide satellites with them in sequence for updating $TN$. It is worth noting that, before a new task being assigned to the satellite to make a decision, it will be screened first. If there is a conflict with the current state of the satellite, the task will be skipped. In this way, we can ensure that the final selection sequence of tasks is consistent with the constraints. Also, this method is a form of pruning as demonstrated in Fig.\ref{fig:Pruning}. As shown on picture, if the task has conflicts with the state, the satellite will only have to reject it. So the agent does not need to learn accepting a comflicted task is not a good option, since the environment only provides it with only one option. Tis design reduces the amount of state needed to search and greatly reduces the complexity of calculations.

$L_{TW}$ is a feature label for the remaining time window of the satellite. It is an important element for implementing PPO in this problem. As the satellite decides to accept a task during the learning process, the satellite's remaining time window information will change. Since each window requires two variables to store information and the satellite's time window is continuously divided as shown in Fig.\ref{fig:RTW}, and the required number of variables are also increasing.

\begin{figure}[H]
	\centering  
	\includegraphics[width=0.5\textwidth]{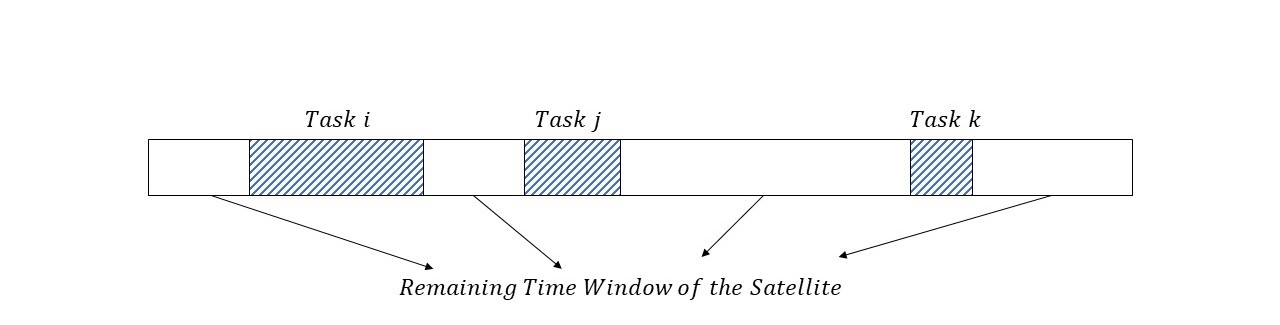}  
	\caption{Remaining Time Window of the Satellite}  
	\label{fig:RTW}   
\end{figure}
Therefore, the time window information cannot be directly applied to the reinforcement learning model based on the neural network, because the neural network cannot handle the dynamic input number. So we propose a feature label to represent the information of the time window. There are many ways to construct $L_{TW}$. Simply, a table can be built to record the time window information of the satellite and $L_{TW}$ can be the corresponding index of the table. Also, time-occupied ratio can be used or more comlicated construction is allowed as long as $L_{TW}$ has a one-to-one match with time window information. In this paper, the label of time window is defined in \ref{LTW}
\begin{equation}
\label{LTW}
L_{TW}=\sum (EndTime_{i}-StartTime_{i})/ScalingFactor
\end{equation}
Scaling Factor is one way to better the performance of the neural networks, which will be discussed later.

$RA$ is the roll angle of the satellite. It is required that when the normal plane of the satellite velocity passes the target, the optical axis of the satellite must point to it and the roll angle at that time is the target angle of that task. What is more, the satellite must maneuver to the target angle before the target point enters the satellite's sensor range for better observation results. Attitude maneuver time is calculated based on the traditional satellite attitude maneuver model, which is basically a process of uniform acceleration, uniform speed, and uniform deceleration.

\section{Mission simulation and results}

\subsection{Initialization of tasks}
In order to verify the effectiveness of the algorithm, a LEO satellite is chosen for its task schedule.Its orbit parameters are shown in Table \ref{OrbitParameter}. The orbit start epoch is 2019/12/30/15:00 UTC. Also, fifty point targets are chosen as the observation targets. The longtitude and latitude of those targets are listed in the appendix and they are also shown in Picture \ref{fig:RTW}  .

Actually, there is no benchmark for the satellite mission task scheduling. The data we use in this paper is only for demonstration. Fifty targets are conbined with pre-set reward and scheduled in thirty minutes.  Rewards of those targets are set randomly as shown in Table \ref{table_example}. But there is also no benchmark for setting the rewards,, which makes it hard to demonstrate the performance of the method since different rewards in reinforcement learning can result in much various situations.

\begin{table}[htbp]

 \caption{Orbit Parameters}
  \label{OrbitParameter}
 \setlength{\tabcolsep}{7mm}
 \begin{center}
 \begin{tabular}{ccc}
  \toprule
Semiaxis & Eccentricity & Inclination\\
  \midrule
6800 km & 0.01 & 55(deg)  \\
  \midrule
 RAAN & Perigee & TA \\
 \midrule
 90(deg) & 90(deg) & 0(deg) \\
  \bottomrule
 \end{tabular}
 \end{center}
\end{table}

\begin{figure}[H]
	\centering  
	\includegraphics[width=0.5\textwidth]{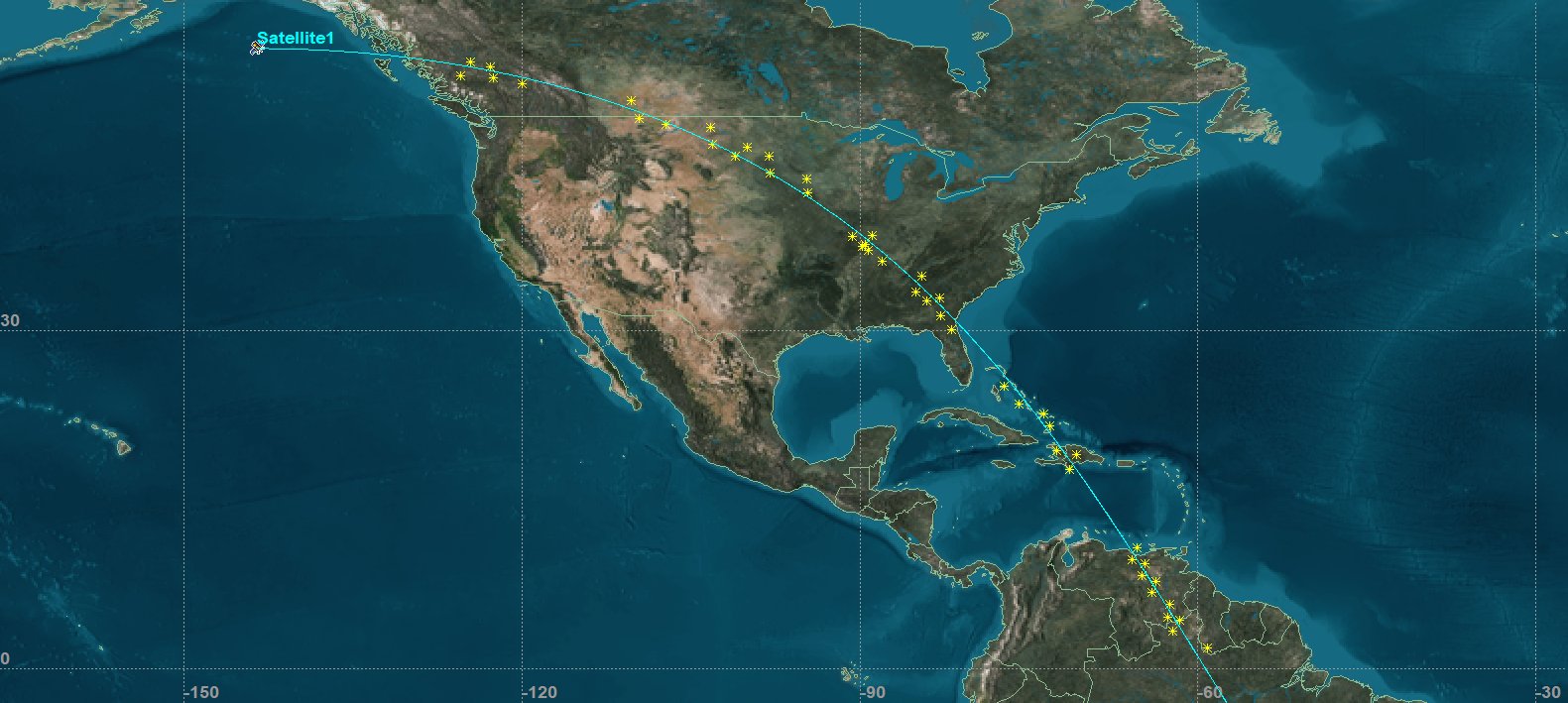}  
	\caption{Demonstration}  
	\label{fig:RTW}   
\end{figure}

\subsection{Performance analysis}

We run the program for several times since PPO algorithm contains random factors which will result in different solutions. Four results are chosen to be demonstrated in this paper. For each time we run the program, 80 times'  training are executed and the solution we get in every running is the one who has the highest reward. Those graphs show the training process of every execution.

\begin{figure}[H]
	\centering  
	\includegraphics[width=0.5\textwidth]{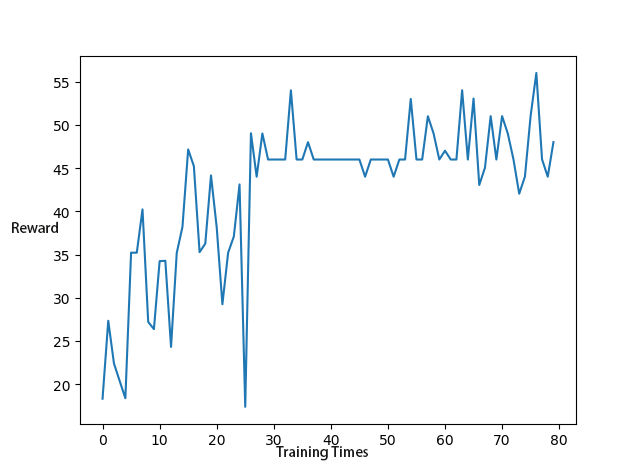}  
	\caption{Result 1}  
	\label{fig:RTW}   
\end{figure}

\begin{figure}[H]
	\centering  
	\includegraphics[width=0.5\textwidth]{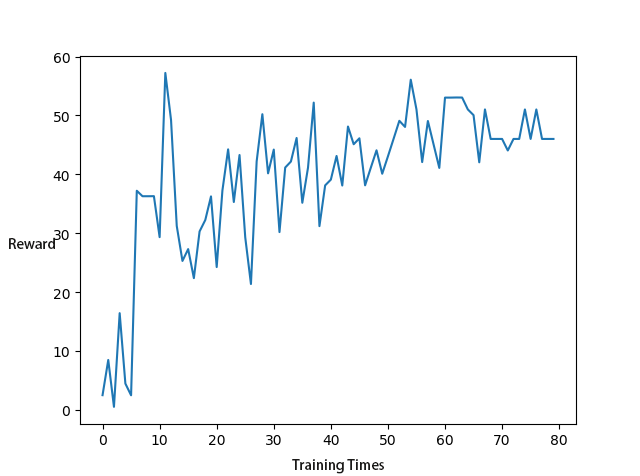}  
	\caption{Result  2}  
	\label{fig:RTW}   
\end{figure}

\begin{figure}[H]
	\centering  
	\includegraphics[width=0.5\textwidth]{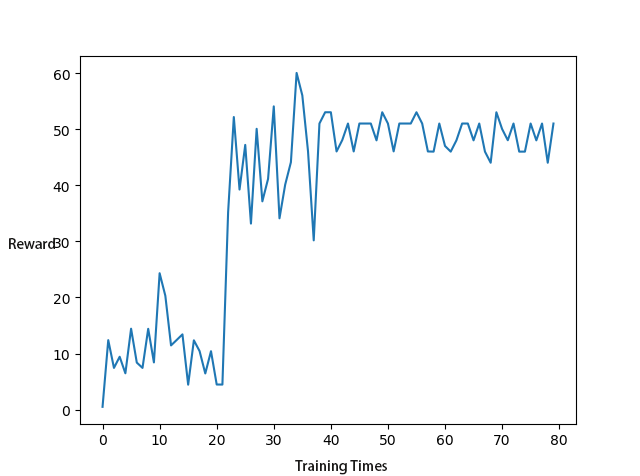}  
	\caption{Result 3}  
	\label{fig:RTW}   
\end{figure}

\begin{figure}[H]
	\centering  
	\includegraphics[width=0.5\textwidth]{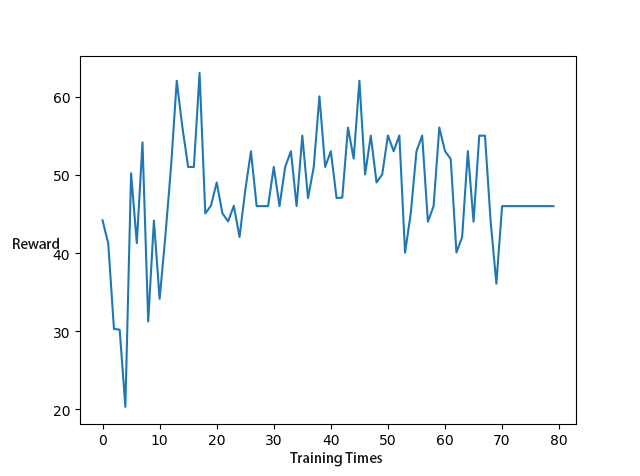}  
	\caption{Result 4}  
	\label{fig:RTW}   
\end{figure}

In each running, we can get differents solutions. They are shown in Table \ref{Solution}. By running it for four times,  a solution with 63 is obtained.  It is hard to say that whether it is a good result or not since the reward and the attributes of those targets are chosen artifiicially. Those presettings will have an influence on the the performance of different algorithm for that lots of algorithms have their own property which mean more proper to schedule tasks with special attributes like low level of conflicts. We hope this propblem can be solved in the future.

\subsection{Comparison}
 We just use FCFS(First come first serve)  algorithm to be compared with our PPO algorithm. 
It  chooses Task 1, 7, 10, 12, 13, 15, 18, 20, 21, 25, 27, 30, 32, 33, 34, 36, 38, 40, 44, 46, 48, 50
and the corresponding reward is 46, which is much smaller than the PPO algorithm. We do not do other comparisons here and more things are waited to be discussed here. We realize this is not a convincing method to demonstrate the performance of our model and hope more progresses can be archieved in the future.

\section{Conclusion and future work}
This paper builds a basic framework of reinforcement learning's application in satellites' mission schedule. In this paper, PPO algorithm is applied and demonstrate its capability in the schedule problem. But more things are need to be discussed. The future work could focus on the improvement of algorithm's stability. More constraints and multi-satellites situation should also be considered in the following work. 

\section*{Appendix}

\appendices

\begin{table}[H]
\renewcommand{\arraystretch}{1.3}
\caption{Positions of 50 targets}
\label{table_example}
\centering
\setlength{\tabcolsep}{4mm}
\begin{tabular}{cccc}
\hline
Number & Reward &Latitude & Longtitude\\
\hline
1 &2& 52.608 &	-125.448 \\
2 & 7&53.862 &	-124.534 \\
3 &4& 53.472 &	-122.831 \\
4 & 4&52.447 &	-122.543 \\
5 & 2&51.913 &	-119.985 \\
6 & 5&52.075 &	-116.494 \\
7 &2& 50.851 &	-116.014 \\
8 & 3&50.961 &	-112.739 \\
9 & 3&50.385 &	-113.687 \\
10 &2& 50.426 &	-110.341 \\
11 &5& 48.893 &	-109.618 \\
12 &2 &48.28 &	-107.243 \\
13 &2& 48.063 &	-103.296 \\
14 &4& 46.556 &	-103.078 \\
15 &2& 46.328 &	-100.019 \\
16 &3& 45.506 &	-101.08 \\
17 &4 &45.504 &	-98.095 \\
18 &2& 44.014 &	-97.972 \\
19 &4 &43.461 &	-94.685 \\
20 &2& 42.237 &	-94.648 \\
21 &2& 38.389 &	-90.711 \\

\end{tabular}
\end{table}

\begin{table}[H]
\renewcommand{\arraystretch}{1.3}
\setlength{\tabcolsep}{7mm}
\centering
\begin{tabular}{cccc}
22 & 4&37.456 &	-89.778 \\
23 &3& 38.492 &	-88.948 \\
24 &6& 37.663 &	-89.57 \\
25 &2& 37.145 &	-89.259 \\
26 &7 &36.212 &	-88.015 \\
27 &2& 34.864 &	-84.491 \\
28 &4& 33.413 &	-85.009 \\
29 &4& 32.688 &	-84.076 \\
30 &2& 32.895 &	-82.936 \\
31 &5& 31.34	 & -82.833 \\
32 &2& 30.096 &	-81.9 \\
33 &2& 25.121 &	-77.235 \\
34 &2& 23.462 &	-75.888 \\
35 &4& 22.633 &	-73.711 \\
36 &2& 21.596 &	-73.193 \\
37 &6& 19.316 &	-72.571 \\
38 &2& 19.006 &	-70.808 \\
39 &4& 17.657 &	-71.43 \\
40 &2& 10.712 &	-65.418 \\
41 &4& 9.676 &	-65.833 \\
42 &5& 9.365 &	-64.693 \\
43 &3& 8.225 &	-65.004 \\
44 &2& 7.706 &	-63.76 \\
45 &5& 6.773 &	-64.071 \\
46 &2& 5.737 &	-62.516 \\
47 &5& 4.597 &	-62.723 \\
48 &2& 4.286 &	-61.583 \\
49 &3& 3.353 &	-62.205 \\
50 &2& 1.798 &	-59.199 \\
\hline
\end{tabular}
\end{table}

\begin{table*}[!h]

 \caption{Solutions}
  \label{Solution}
 \setlength{\tabcolsep}{7mm}
 \begin{center}
 \begin{tabular}{ccc}
  \toprule
No. & Chosen tasks & Max rewards \\                         
\midrule
The solution 1 & 1, 5, 7, 10, 12, 13, 16, 17, 19, 21, 26, 27, 30, 32, 33, 34, 36, 38, 40, 44, 46, 48, 50 & 56\\
  \midrule
The solution 2 & 8, 11, 18, 20, 21, 26, 29, 31, 35, 37, 39, 41, 45,48, 50  &57\\
  \midrule
The solution 3 & 3, 6, 8, 11, 14, 16, 17, 19, 21, 25, 27, 30, 32, 33, 34, 36, 38, 40, 44, 46, 48, 50 &60 \\
 \midrule
The solution 4 & 2, 6, 8, 11, 14, 16, 17,19, 21, 25, 27, 30, 32, 33, 34, 36, 38, 40, 44, 46, 48, 50 & 63 \\
  \bottomrule
 \end{tabular}
 \end{center}
\end{table*}

\bibliographystyle{IEEEtran}      

\bibliography{MSP}

\end{document}